\documentclass[10pt,twocolumn,letterpaper]{article}

\usepackage[pagenumbers]{cvpr} 

\usepackage{graphicx}
\usepackage{amsmath}
\usepackage{amssymb}
\usepackage{booktabs}

\usepackage{verbatim}
\usepackage{multirow}
\usepackage{floatrow}
\usepackage[table,xcdraw]{xcolor}
\usepackage{colortbl}
\newfloatcommand{capbtabbox}{table}[][\FBwidth]
\newcommand{\cc}{\cellcolor{orange!20!}}

\usepackage[pagebackref,breaklinks,colorlinks]{hyperref}

\usepackage[capitalize]{cleveref}
\crefname{section}{Sec.}{Secs.}
\Crefname{section}{Section}{Sections}
\Crefname{table}{Table}{Tables}
\crefname{table}{Tab.}{Tabs.}

\def\cvprPaperID{9210} 
\def\confName{CVPR}
\def\confYear{2022}

\begin{document}

\title{TVConv: Efficient Translation Variant Convolution\\ for Layout-aware Visual Processing}

\author{Jierun Chen\textsuperscript{1}, Tianlang He\textsuperscript{1}, Weipeng Zhuo\textsuperscript{1}, Li Ma\textsuperscript{1}, Sangtae Ha\textsuperscript{2}, S.-H. Gary Chan\textsuperscript{1}\\
\textsuperscript{1}The Hong Kong University of Science and Technology,
\textsuperscript{2}University of Colorado at Boulder\\
{\tt\small \{jcheneh, theaf, wzhuo, lmaag, gchan\}@cse.ust.hk, sangtae.ha@colorado.edu }
}
\maketitle


\begin{abstract}
As convolution has empowered many smart applications, dynamic convolution further equips it with the ability to adapt to diverse inputs. However, the static and dynamic convolutions are either layout-agnostic or computation-heavy, making it inappropriate for layout-specific applications, \eg, face recognition and medical image segmentation. We observe that these applications naturally exhibit the characteristics of large intra-image (spatial) variance and small cross-image variance. This observation motivates our efficient translation variant convolution (TVConv) for layout-aware visual processing. Technically, TVConv is composed of affinity maps and a weight-generating block. While affinity maps depict pixel-paired relationships gracefully, the weight-generating block can be explicitly over-parameterized for better training while maintaining efficient inference. Although conceptually simple, TVConv significantly improves the efficiency of the convolution and can be readily plugged into various network architectures. Extensive experiments on face recognition show that TVConv reduces the computational cost by up to $3.1\times$ and improves the corresponding throughput by $2.3\times$ while maintaining a high accuracy compared to the depthwise convolution. Moreover, for the same computation cost, we boost the mean accuracy by up to 4.21\%. We also conduct experiments on the optic disc/cup segmentation task and obtain better generalization performance, which helps mitigate the critical data scarcity issue. Code is available at \url{https://github.com/JierunChen/TVConv}.
\end{abstract}
\section{Introduction}
\label{sec:intro}

With the breakthrough of deep neural networks, we are witnessing a flourishing growth in AI-powered applications and services. While a performance gain usually comes with an overhead of model size and computation, more interest has been devoted to the lightweight and computation-efficient network design, which can unleash the potential for on-device inference for user experience and privacy.

\begin{figure}
  \centering
   \includegraphics[width=1\linewidth]{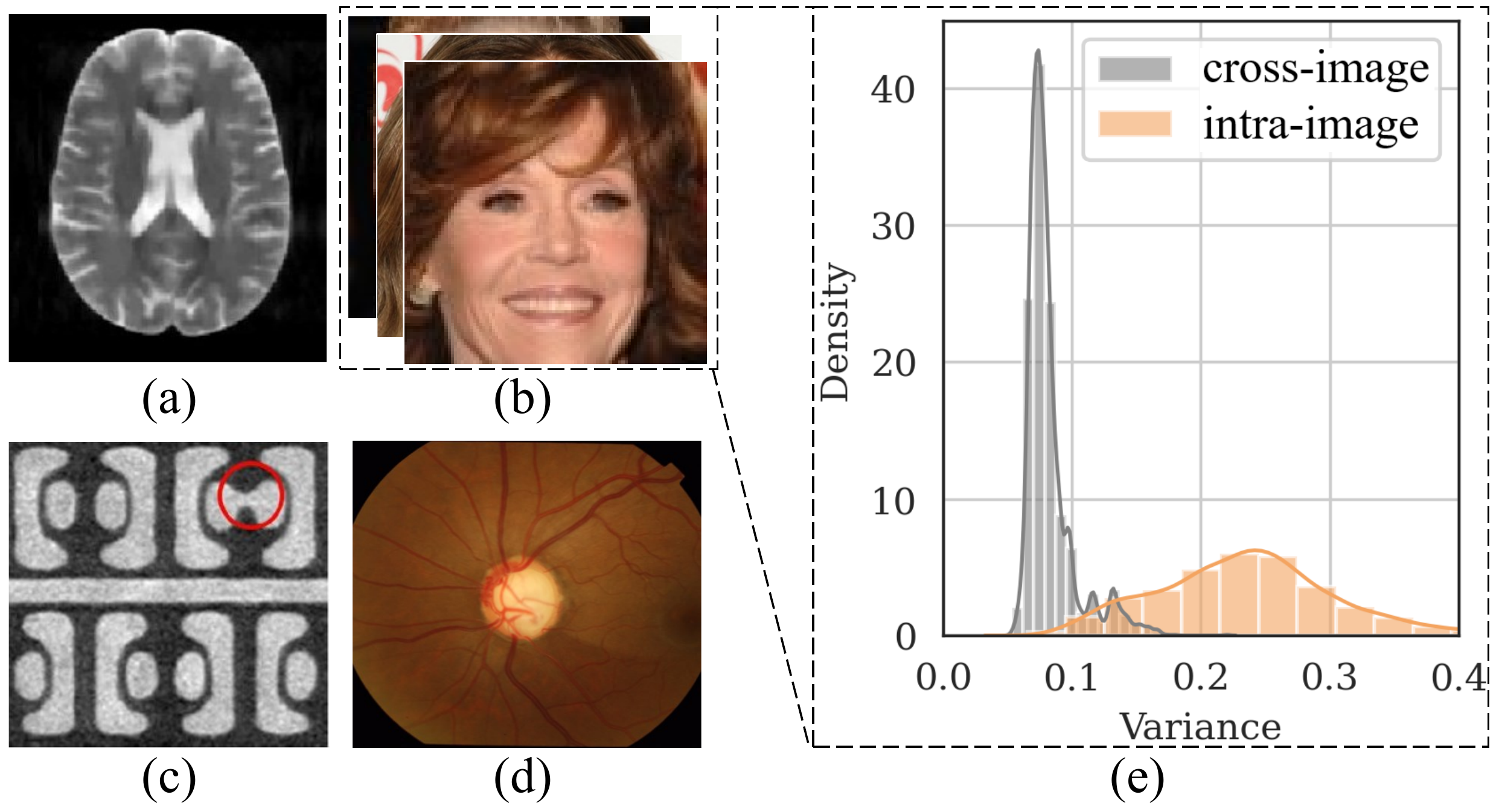}
   \caption{Various applications with specific layout: (a) brain MRI analysis, (b) face recognition, (c) industrial product defects detection and (d) optic disc/cup segmentation. Each application exhibits large intra-image variance and small cross-image variance, as shown in subfigure (e), where the statistic is calculated from the intermediate VGG~\cite{simonyan2014very} feature maps by feeding the LFW face verification dataset~\cite{huang2008labeled}.
   }
   \label{fig:application}
   \vspace{-0.1in}
\end{figure}

Despite various neural network architectures~\cite{sandler2018mobilenetv2,ma2018shufflenet,tan2019efficientnet,he2022tackling} being proposed for efficiency, their fundamental operators remain more or less the same, \eg, vanilla convolution (conv) and depthwise conv. These operators share one key characteristic, \textit{translation equivariance}~\cite{worrall2017harmonic}, \ie, filters are shared spatially in a sliding window manner. Though it saves parameters for a lightweight model, it deprives the model of being adaptive to different positions in an image. Therefore, it has to exhaustively learn many filters for feature matching~\cite{wang2021adaptive}, which results in a massive waste of computation for many tasks with a specific layout.

For layout-specific tasks, as in \cref{fig:application}, the inputs demonstrate a regional statistic with large intra-image (spatial) variance and small cross-image variance. For example, when we use face ID to securely and conveniently unlock our mobile phones, our hair generally appears on the upper region, vertically followed by our forehead, eyes, nose, mouth, jaw, etc. Similar tasks include, but are not limited to, talking head generation, industrial product defects detection, and medical image processing.

To account for the large intra-image variance, many works propose per-pixel dynamic conv~\cite{su2019pixel,wang2021adaptive,li2021involution,zhou2021decoupled,chen2021dynamic,jin2021laconv,wang2019carafe}, which extend the per-image dynamic conv~\cite{yang2019condconv,jia2016dynamic,chen2020dynamic,ma2020weightnet,zhang2020dynet} to a spatial domain. They apply the local feature dependant filters by assembling multiple templates for each position. However, it can easily suffer from prohibitively large memory footprints and substantial computation overhead. Besides, it omits the nature of small cross-image variance for layout-specific tasks, causing redundant computation of pixel-wise filters for a series of the inputs.
Therefore, the challenge remains to exploit an efficient operator to better serve such applications. 

\begin{figure}
  \centering
   \includegraphics[width=1\linewidth]{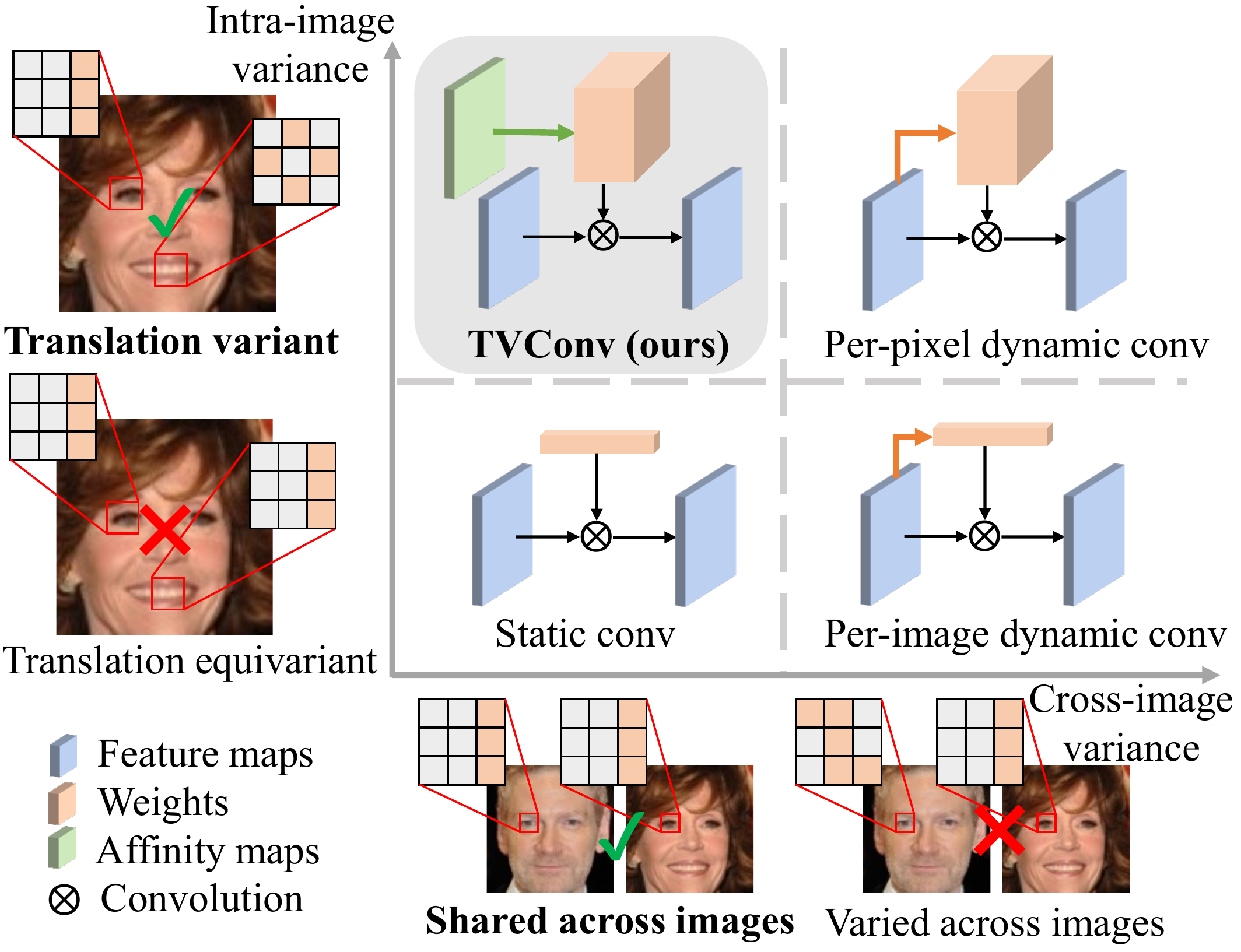}
   \caption{Comparison among conv variants. Different from the static conv, per-image dynamic conv and per-pixel dynamic conv, our TVConv is translation variant and shared across images, superbly fitting into the layout-specific applications.
   }
   \label{fig:scheme_comparison}
   \vspace{-0.1in}
\end{figure}

In this paper, we propose an efficient fundamental operator called Translation Variant Convolution (TVConv) for layout-aware visual processing. In contrast to existing conv variants, TVConv elegantly supports the applications with a specific layout by its \textit{translation variance} and the property of being \textit{shared across images}, as in \cref{fig:scheme_comparison}.  Technically, we first formulate compact and learnable affinity maps to distinguish various local features. The affinity maps implicitly capture the semantic relationships among different regions, similar to the attention mechanism~\cite{vaswani2017attention,woo2018cbam}, but without bothering to derive a heavy affinity matrix. We then feed the affinity maps into a weight-generating block to yield the weights, which are then applied as filters on the input. Unlike the dynamic conv, whose weight-generating block is constrained by the tradeoff between the adaptivity and the computation overhead, TVConv can be freely over-parameterized to strengthen the spatial adaptivity without slowing down the inference speed. The affinity maps are fixed once they are trained. Thus the weight generating process can be performed only once during initialization. Then it produces and caches the weights in memory, allowing them to be fetched efficiently for subsequent inferences. Through extensive experiments on face recognition, TVConv is shown to strike a better tradeoff between accuracy and computation complexity. Simply replacing the prevalent depthwise conv in various architectures (\eg MobileNetV2, ShuffleNetV2), TVConv reduces the theoretical complexity by up to $\mathbf{3.1\times}$ and correspondingly accelerates the throughput by $\mathbf{2.3\times}$ while maintaining a high accuracy. On the other hand, TVConv boosts the mean accuracy by up to 4.21\% under an extreme low complexity constraint. Moreover, thanks to the layout awareness, TVConv shows a better generalization ability for optic disc/cup segmentation on unseen datasets, which helps to alleviate the data-scarcity dilemma for medical image analysis.

In short, our contributions include: (1) We rethink the existing convolution variants of their inappropriate properties for layout-specific tasks in terms of the intra-image and cross-image variance; (2) By being translation variant and shared across images, an efficient fundamental operator, TVConv, is proposed for layout-aware visual processing; (3) Extensive experiments show that TVConv clearly improves the efficiency for face recognition and generalize better for medical image processing.

\section{Related Work}
\label{sec:related_work}
This section briefly reviews related works considering the macro network design to the micro operators, such as the dynamic conv and the attention mechanism.

\medskip\noindent\textbf{Efficient network design.}
Stimulated by the practical needs of edge intelligence, an increasing interest has been put into the field of efficient network design. MobileNets~\cite{howard2017mobilenets, sandler2018mobilenetv2} are built principally on the depthwise separable conv~\cite{sifre2014rigid}, which works by decomposing a standard conv into a depthwise and a pointwise conv. ShuffleNets~\cite{ma2018shufflenet,zhang2018shufflenet} further shuffle the channels to facilitate the information flow. Later on, supported by the neural architecture search~\cite{zoph2016neural}, MobileNetV3~\cite{howard2019searching}, MnasNet~\cite{tan2019mnasnet} and EfficientNets~\cite{tan2019efficientnet,tan2021efficientnetv2} employ reinforcement learning to search for efficient networks. Despite various architectures being proposed, one of their fundamental operators remains primarily a depthwise conv. Our TVConv inherits its lightweight merits and takes one step further to be translation variant and layout-aware. 

\medskip\noindent\textbf{Dynamic conv.}
Instead of blindly stacking more static convs to increase models' capacity, dynamic conv leads a new wave to apply adaptive filters conditional on the input. Recent works~\cite{yang2019condconv,ma2020weightnet,zhang2020dynet,wang2020dynamic} attempt to use per-image dependant filters, either by discrete gating or continuously weighted averaging of multiple templates. Another branch of works~\cite{wang2019carafe,zhou2021decoupled,chen2021dynamic,li2021involution} further extends the adaptiveness to the spatial dimension by using per-pixel dynamic conv. 
A pioneering work~\cite{jia2016dynamic} demonstrates its effectiveness on video and stereo prediction. Deformable conv~\cite{dai2017deformable,zhu2019deformable} augments the filter shape with auxiliary offsets. Some works~\cite{mildenhall2018burst,xu2020unified} apply the dynamic conv to real-world image restoration. Our method follows the content-aware paradigm but is more efficient for layout-specific tasks. As the tasks exhibit a property of small cross-class variance, TVConv is designed to be shared across images and exempted from the heavy and redundant filter recalculation.

\medskip\noindent\textbf{Attention mechanism.}
Attention mechanism originates from the field of machine translation~\cite{vaswani2017attention}. It excels at capturing long-range dependency by summarizing the global context and reweighting responses at each location. Its tremendous success in the language domain has motivated researches to explore the applicability to computer vision, including image generation~\cite{zhang2019self,parmar2018image}, object detection~\cite{wang2018non,hu2018relation} and semantic segmentation~\cite{fu2019dual,huang2019ccnet}. While some works~\cite{park2018bam,woo2018cbam} propose attention module as a versatile and orthogonal supplement to existing conv operator, more recent works \cite{dosovitskiy2020image,liu2021swin} aggressively adopt purely attention-powered architectures. Although significant improvements have been achieved, the above-mentioned attention mechanism involves a heavy computation of the affinity matrix that scales quadratically with the resolution of the input. In contrast, the affinity maps of our TVConv preserve the appealing ability to attend the whole image and obtain pixel-paired relationships, but more efficiently.
\section{Approach}
\label{sec:approach}
In this section, we start with a conventional conv and then describe how it grows to our TVConv for layout-aware processing. After detailing the implementation, we elaborate its connections with prior operators. 
\subsection{Preliminary}
The depthwise conv has been widely adopted in state-of-the-art light-weight architectures. Given an input tensor 
$ \mathbf{I} \in \mathbb{R}^{ c \times h \times w} $
, it computes the output 
$ \mathbf{O} \in \mathbb{R}^{ c \times h \times w} $ 
by channel-wisely convolving a local patch 
$ \mathbf{P}_{i,j} \in \mathbb{R}^{ c \times k \times k} $
with the spatially shared weight 
$ W \in \mathbb{R}^{ c \times k \times k} $,
where 
$ (i,j) $ 
denotes a spatial location
and 
$ \mathbf{P}_{i,j} $ 
is centered at 
$ \mathbf{I}_{i,j} \in \mathbb{R}^{ c } $ 
with 
$ k \times k $ 
size.
For conciseness, we omit the bias term and formulate the process as follows:
\begin{equation}
  \mathbf{O}_{i,j} = W \otimes \mathbf{P}_{i,j},
  \label{eq:depthwise}
\end{equation}
where 
$ \otimes $ 
denotes a channel-wise convolution. As the output 
$ \mathbf{O} $ 
shares the same number of channels 
$ c $ 
with the input 
$ \mathbf{I} $,
a depthwise conv is normally followed by a pointwise
($ 1 \times 1 $)
conv for channel projection and fusion.
\subsection{Design of TVConv}
The depthwise conv shares the same weight across the image. This translation equivariance makes it layout-agnostic and inefficient for layout-specific tasks. In contrast, one may attempt to revert this property to be translation variant:
\begin{equation}
  \mathbf{O}_{i,j} = \mathbf{W}_{i,j} \otimes \mathbf{P}_{i,j}.
  \label{eq:naive_pixelwise}
\end{equation}
This, however, results in a bulky weight tensor 
$ \mathbf{W} \in \mathbb{R}^{ c \times k \times k \times h \times w} $.  
The unaffordable number of parameters impedes efficient training and may be prone to overfitting. Thus, we propose to first decompose 
$ \mathbf{W} $ as:
\begin{equation}
  \mathbf{W'} = \mathbf{B'} \mathbf{A'},
  \label{eq:matrix_factorization}
\end{equation}
where 
$ \mathbf{W'} \in \mathbb{R}^{ ( c \times k \times k ) \times ( h \times w )} $ 
is the reshaped version of 
$ \mathbf{W} $, 
$ \mathbf{B'} \in \mathbb{R}^{ ( c \times k \times k ) \times c_A  } $ 
is the basis matrix, and 
$ \mathbf{A'} \in \mathbb{R}^{ c_A \times ( h \times w )} $ 
is the coefficient matrix. By doing so, the number of parameters is greatly reduced from 
$ ( c k k h w ) $ 
to
$ ( c  k  k  c_A + c_A  h w ) $.
In general, $ c_A $ 
can be set to a small value, \eg, $c_A = 1$,
approximately yielding a 
$ ( c k k h w ) / ( c k k + h w ) \approx  ( c k k ) $
times reduction of parameters. 

To further strengthen the translation variance, we replace the linear multiplication in \cref{eq:matrix_factorization} by a non-linear function:
\begin{equation}
  \mathbf{W} = \mathcal{B} (\mathbf{A}),
  \label{eq:translation_variant}
\end{equation}
where
$ \mathbf{A} \in \mathbb{R}^{ c_A \times h \times w } $  
denotes our affinity maps and 
$ \mathcal{B} $ 
is a non-linear function, instantiated as our weight generating block. For the affinity maps 
$ \mathbf{A} $, 
they gracefully depict the pixel-paired relationships in a compact size. For the weight-generating block 
$ \mathcal{B} $, 
it perceives the affinity maps and produces the weights for operation.
\begin{figure}
  \centering
   \includegraphics[width=1\linewidth]{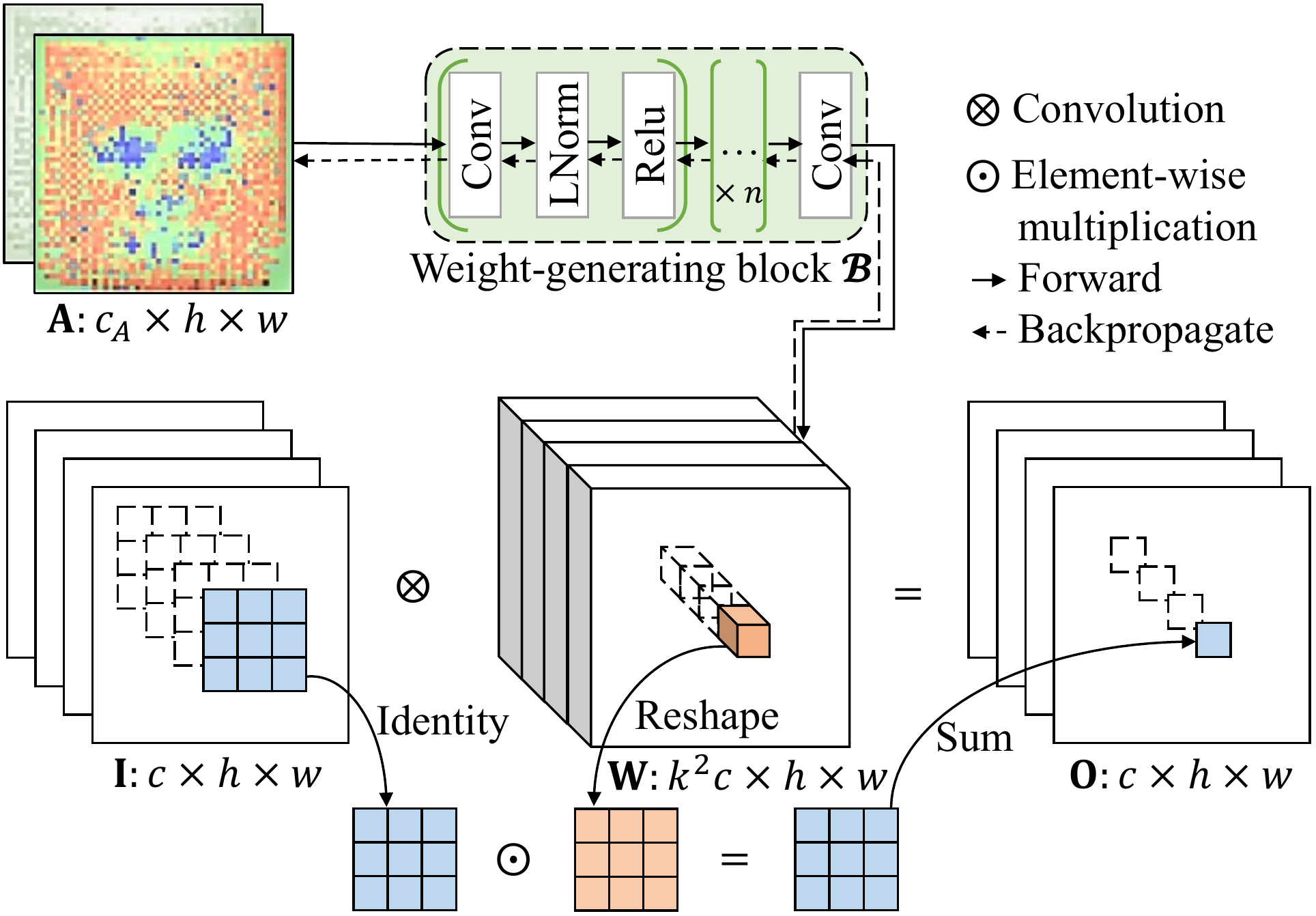}
   \caption{Schematic illustration of our proposed TVConv. The learnable affinity maps $ \mathbf{A} \in \mathbb{R}^{ c_A \times h \times w} $ are fed into a weight-generating block $ \mathcal{B} $ to yield the weight $ \mathbf{W} \in \mathbb{R}^{ k^2c \times h \times w} $ (reshaped version) for depthwise conv. Each patch within the input feature maps $ \mathbf{I} \in \mathbb{R}^{ c \times h \times w} $ is then element-wisely multiplied by the corresponding weight patch within $ \mathbf{W} $, and followed by a summation to produce each value within the output feature maps $ \mathbf{O} \in \mathbb{R}^{ c \times h \times w} $.}
   \label{fig:framework}
   \vspace{-0.05in}
\end{figure}

Thanks to the conciseness of our design, TVConv can be fast prototyped, as illustrated in~\cref{fig:framework}. There are two processes involved: generating and applying the weights. While the latter process simply mirrors the depthwise conv, the weight generation is the focus of our work. The weight generating block $ \mathcal{B} $ takes in the affinity maps $ \mathbf{A} $ and feeds through a standard conv, a layer normalization, and an activation (\eg Relu). These three layers can be consecutively executed multiple times and finally followed by an output layer. Unlike most other convolutional neural networks which use the batch normalization~\cite{ioffe2015batch}, we adopt the layer normalization~\cite{ba2016layer} within $ \mathcal{B} $ because the affinity maps $ \mathbf{A} $ are shared for different inputs and the ``batch size'' for $ \mathbf{A} $ can be 
interpreted as 1. The affinity maps $ \mathbf{A} $  can be trained end-to-end by standard back-propagation since all the operations involved are differentiable.

As shown in the work~\cite{allen2019convergence}, over-parameterization (\eg increase network width) helps the training and achieves better performance. However, it sacrifices the test-time speed. Notably, our TVConv inherits its advantages while avoiding the computational overhead. TVConv allows the weight generating block $ \mathcal{B} $ to be over-parameterized up to the memory limit. After training, the affinity maps $ \mathbf{A} $ are fixed. Therefore, we can perform the weight generation process only once during initialization and achieve fast inference by readily applying the same weights. 

As a versatile replacement of depthwise conv, TVConv can be easily plugged into the bottleneck blocks of various architectures (\eg MobiletNets, ShuffleNets, MnasNet). These networks provide a throttleable hyper-parameter, the width of the network. Thanks to the layout awareness, TVConv can adaptively perceive different regions and allow the width to be greatly narrowed down while maintaining the performance.

\subsection{Connection with prior operators}
To further distinguish TVConv from prior operators, here we discuss the connections between TVConv with the static/dynamic conv and the self-attention.

\medskip\noindent\textbf{Static conv.} The diversity within affinity maps $ \mathbf{A} $ is the key to our TVConv, which implicitly measures the potential gap between TVConv with the static translation equivariant conv. When the learned affinity maps are filled with a constant value, TVConv degrades to the static depthwise conv, as $ \mathbf{W}_{i,j} = W $ for different $ (i,j) $. We visualize and discuss the learned affinity maps in \cref{sec:experiments}.

\medskip\noindent\textbf{Dynamic conv.} Formally by replacing the affinity maps $ \mathbf{A} $ in \cref{eq:translation_variant} with the input $ \mathbf{I} $, TVConv would fall into the category of dynamic conv. As a result, the weight generating process has to be repeated for different inputs, and the weight generating block has to be carefully kept compact. Besides, though the weights generated by dynamic conv is translation variant, we clarify that the dynamic conv itself is translation equivariant~\cite{wang2021adaptive}. The dynamic conv cannot predict the adaptive weights before ``seeing'' the input. Therefore, dynamic conv is layout-adaptive but not layout-aware.

\medskip\noindent\textbf{Self-attention.}
Our work is also related to the self-attention \cite{vaswani2017attention}. Its core idea is to aggregate the global context with adaptive weights.
This complicated operator has two important characters: similar inputs in different positions would respond similarly (regardless of the position encoding); response at each position is aware of the global context. Intriguingly, our affinity maps $ \mathbf{A}$ resemble these two characters implicitly. Firstly, if two positions within $ \mathbf{A}$ appear with similar values, they would produce similar weights for computing. Secondly, once the affinity maps are learned from the data, the pixel-paired relationships are formulated and fixed within the spatial domain, making them mutually context-aware. 
In short, while preserving the appearing properties of self-attention, our affinity maps are yet much more efficient for layout-specific applications.

\section{Experiments}
\label{sec:experiments}
In this section, we carry out a systematical evaluation of TVConv and report the experimental results. We start with face recognition to validate its efficiency. We then move to the optic disc/cup segmentation task, suggesting a promisingly better generalization by applying the TVConv. We also provide a comprehensive ablation study to consolidate various considerations for implementation.

\begin{figure*}
\begin{floatrow}
\ffigbox[0.29\textwidth]{%
  \includegraphics[width=0.26\textwidth]{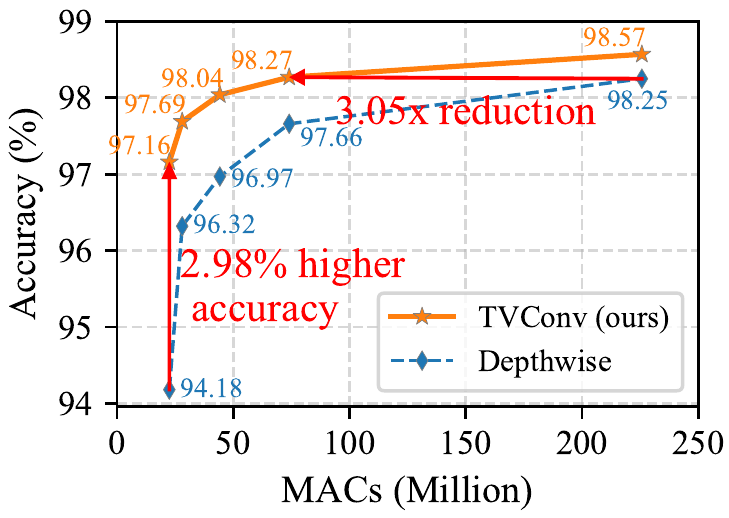}
  {%
    \vspace{-0.03in}%
    \subcaption{MobileNetV2 as the architecture}%
  }
  
  \includegraphics[width=0.26\textwidth]{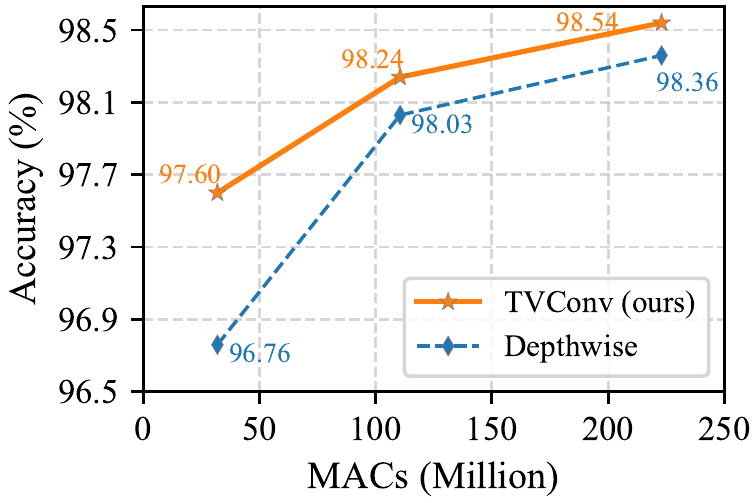}
  {%
    \vspace{-0.03in}%
    \subcaption{ShuffleNetV2 as the architecture}%
  }
}{%
  \caption{The accuracy-complexity envelope on the LFW dataset.}%
  \label{fig:acc_comp_envelope}%
  \vspace{-0.1in}%
}
\capbtabbox{%
\resizebox{0.65\textwidth}{!}{%
\begin{tabular}{@{}clclllll@{}}
\toprule
  \multicolumn{1}{c}{\multirow{2}{*}{\begin{tabular}[c]{@{}c@{}}Arch× \\ width\end{tabular}}} &
  \multicolumn{1}{c}{\multirow{2}{*}{\begin{tabular}[c]{@{}c@{}}MACs \\ (M)\end{tabular}}} &
  \multirow{2}{*}{Operator} &
  \multicolumn{5}{c}{Accuracy across datasets (\%)} \\ \cmidrule(l){4-8} &
  \multicolumn{1}{c}{} &
   &
  \multicolumn{1}{c}{LFW} &
  \multicolumn{1}{c}{CFP-FF} &
  \multicolumn{1}{c}{AgeDB-30} &
  \multicolumn{1}{c}{CALFW} &
  \multicolumn{1}{c}{Mean} \\ \midrule
\multirow{2}{*}{MB×0.1} & \multirow{2}{*}{22.47} & Depthwise   & 94.18 ± 0.34      & 93.01 ± 0.40      & 79.10 ± 0.44      & 82.50 ± 0.59      & 87.20        \\
                      &                          & \cc{TVConv} & \cc{97.16 ± 0.11} & \cc{96.49 ± 0.21} & \cc{84.61 ± 0.20} & \cc{87.37 ± 0.21} & \cc{91.41}   \\
\multirow{2}{*}{MB×0.2} & \multirow{2}{*}{28.00} & Depthwise   & 96.32 ± 0.07      & 95.53 ± 0.26      & 82.56 ± 0.32      & 85.62 ± 0.31      & 90.01        \\
                      &                          & \cc{TVConv} & \cc{97.69 ± 0.12} & \cc{97.21 ± 0.26} & \cc{85.97 ± 0.38} & \cc{87.96 ± 0.38} & \cc{92.21}   \\
\multirow{2}{*}{MB×0.3} & \multirow{2}{*}{44.20} & Depthwise   & 96.97 ± 0.27      & 96.31 ± 0.21      & 84.72 ± 0.17      & 86.87 ± 0.50      & 91.22        \\
                      &                          & \cc{TVConv} & \cc{98.04 ± 0.16} & \cc{97.73 ± 0.19} & \cc{86.99 ± 0.45} & \cc{88.97 ± 0.58} & \cc{92.93}   \\  
\multirow{2}{*}{MB×0.5} & \multirow{2}{*}{74.03} & Depthwise   & 97.66 ± 0.16      & 96.98 ± 0.18      & 85.74 ± 0.46      & 88.08 ± 0.23      & 92.11        \\
                      &                          & \cc{TVConv} & \cc{98.27 ± 0.07} & \cc{97.96 ± 0.19} & \cc{87.88 ± 0.12} & \cc{89.22 ± 0.22} & \cc{93.33}   \\  
\multirow{2}{*}{MB×1.0} & \multirow{2}{*}{225.72}& Depthwise   & 98.25 ± 0.13      & 97.82 ± 0.11      & 88.00 ± 0.21      & 89.41 ± 0.26      & 93.37        \\
                      &                          & \cc{TVConv} & \cc{98.57 ± 0.12} & \cc{98.43 ± 0.09} & \cc{89.58 ± 0.19} & \cc{90.29 ± 0.11} & \cc{94.22}   \\
                      \midrule
\multirow{2}{*}{SF×0.5} & \multirow{2}{*}{31.95} & Depthwise   & 96.76 ± 0.08      & 95.95 ± 0.08      & 83.87 ± 0.37      & 86.71 ± 0.44      & 90.82        \\
                      &                          & \cc{TVConv} & \cc{97.61 ± 0.22} & \cc{96.99 ± 0.11} & \cc{85.86 ± 0.49} & \cc{88.5 ± 0.35}  & \cc{92.24}   \\
\multirow{2}{*}{SF×1.0} & \multirow{2}{*}{110.53}& Depthwise   & 98.03 ± 0.13      & 97.43 ± 0.18      & 86.80 ± 0.43      & 88.75 ± 0.42      & 92.75        \\
                      &                          & \cc{TVConv} & \cc{98.24 ± 0.09} & \cc{97.83 ± 0.19} & \cc{87.83 ± 0.78} & \cc{89.46 ± 0.37} & \cc{93.34}   \\
\multirow{2}{*}{SF×1.5} & \multirow{2}{*}{222.99}& Depthwise   & 98.36 ± 0.12      & 97.73 ± 0.15      & 87.96 ± 0.40      & 89.43 ± 0.51      & 93.37        \\
                      &                          & \cc{TVConv} & \cc{98.54 ± 0.13} & \cc{98.23 ± 0.08} & \cc{88.68 ± 0.50} & \cc{89.91 ± 0.16} & \cc{93.84}   \\  
                      \bottomrule
\end{tabular}%
}

}{%
    \caption{TVConv consistently outperforms the depthwise conv across four face verification datasets under varied network width for both MobileNetV2 (MB) and ShuffleNetV2 (SF).}%
    \label{tab:varied_width}%
}
\vspace{-0.1in}%
\end{floatrow}
\end{figure*}
\begin{figure*}
  \centering
  \includegraphics[width=1\linewidth]{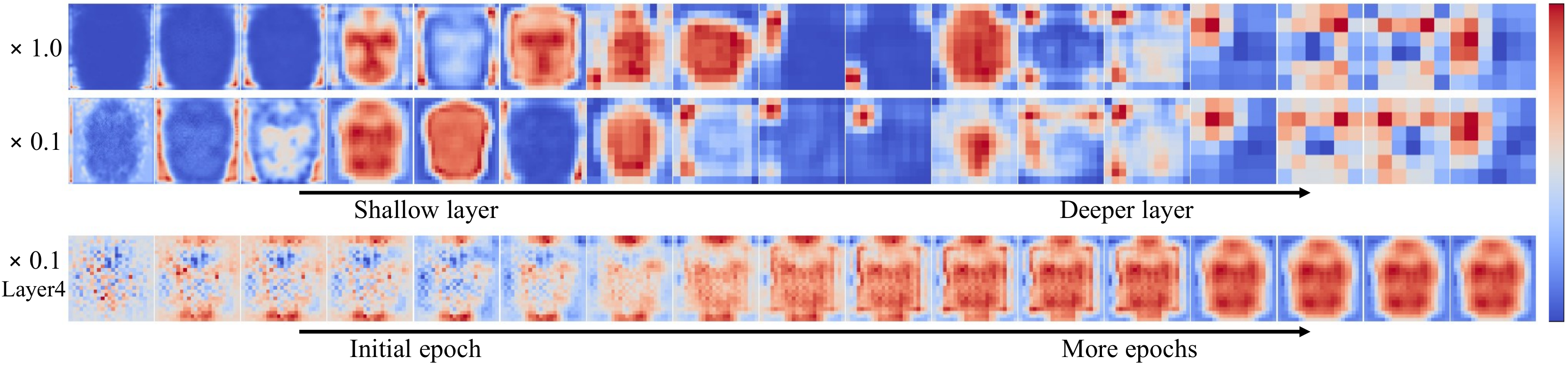}
  \caption{Visualization of our learned affinity maps. The first two rows depict affinity maps in MobileNetV2 x1.0 and x0.1,  as the layer goes deeper from left to right. The bottom row shows the convergence process of an affinity map in MobileNetV2 x0.1 layer 4.
  }
  \label{fig:aff_visualization}
  \vspace{-0.1in}
\end{figure*}

\subsection{Face recognition}\label{face_recognition}
The focus of this work is to provide an efficient operator for layout-aware processing. Therefore, we replace the depthwise conv in various architectures with TVConv. Instead of building upon some bulky models, we use two popular lightweight architectures, MobileNetV2~\cite{sandler2018mobilenetv2} and ShuffleNetV2~\cite{ma2018shufflenet} as baselines (with stride $=1$
for the first layer, following the work~\cite{chen2018mobilefacenets}). EfficientNet~\cite{tan2019efficientnet} is not included because it involves a compound scaling factor associated with width, depth, and resolution, while scaling the width factor is more critical for evaluation in this case. 
For training, we use the publicly available and widely adopted CASIA-WebFace~\cite{yi2014learning} dataset, which consists of 490k images from 10k identities. As it contains many profile view images, we purify it to 329k frontal images by training a simple frontal/profile classifier with the help of the dataset CFP-FP~\cite{sengupta2016frontal}. For validation, as a common practice, we use the face verification datasets LFW~\cite{huang2008labeled}, CFP-FF~\cite{sengupta2016frontal}, AgeDB-30~\cite{moschoglou2017agedb} and CALFW~\cite{zheng2017cross}. The inputs are resized to 96x96 with a horizontal flipping augmentation.

All the models are trained with AM-Softmax~\cite{wang2018additive} loss for training stability. The SGD optimizer is used with momentum 0.9 and weight decay $5\text{e-}4$. The learning rate starts from 0.1 and is divided by 10 at epochs 22, 30, and 35. Total training epochs are 38, with a batch size of 512. We initialize the affinity maps in TVConv with a constant of 1 (refer to \cref{ablation} for further examination). Overall, we measure the accuracy of the validation set relative to computational cost in Multiply-Accumulate operations (MACs), throughput (FPS) and peak memory consumption. The accuracy is reported as the mean and standard deviation over 5 runs if not otherwise stated. We set the batch size as 1/2/4/8/16 and the number of threads as 1/2/4, and report the maximum throughput on a 4GB Raspberry Pi 4B for each model. The peak memory is measured with input size [8, 3, 96, 96].

\begin{figure*}
    \newcommand \AccThrMemwidth{0.245}
    \vspace{-0.12in}
    \centering
        \begin{subfigure}[b]{\AccThrMemwidth \linewidth}
        \centering
        \includegraphics[width=\linewidth]{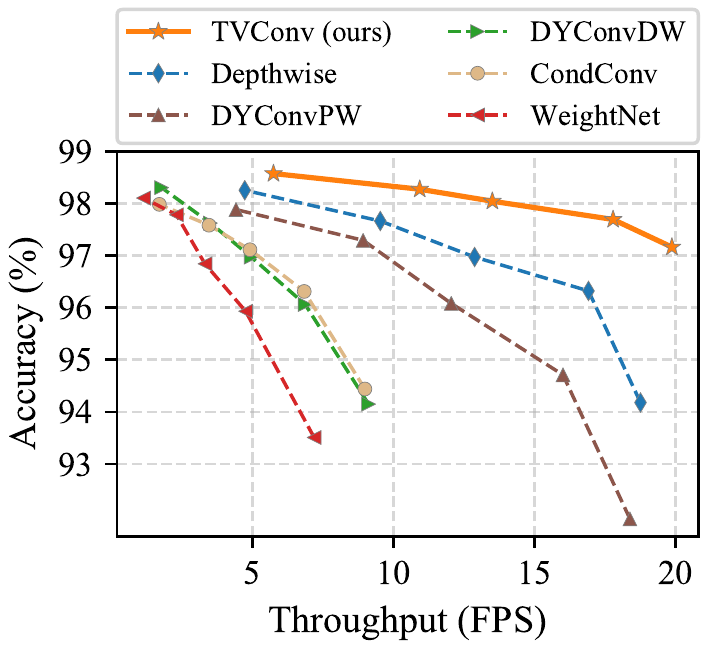}
        \caption{Against \textbf{per-image} dynamic conv}
        \label{fig:AccThrMem_a}
    \end{subfigure}
    \begin{subfigure}[b]{\AccThrMemwidth \linewidth}
        \centering
        \includegraphics[width=\linewidth]{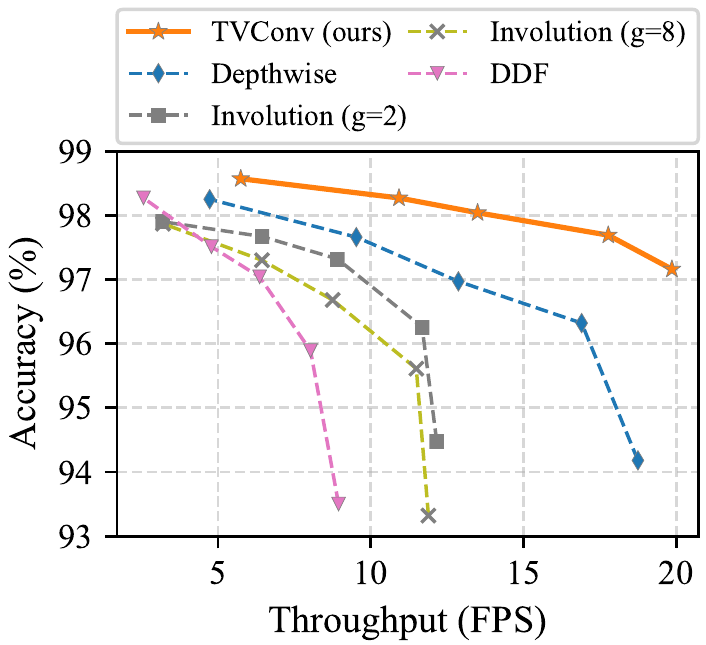}
        \caption{Against \textbf{per-pixel} dynamic conv}
        \label{fig:AccThrMem_b}
    \end{subfigure}
    \begin{subfigure}[b]{\AccThrMemwidth \linewidth}
        \centering
        \includegraphics[width=\linewidth]{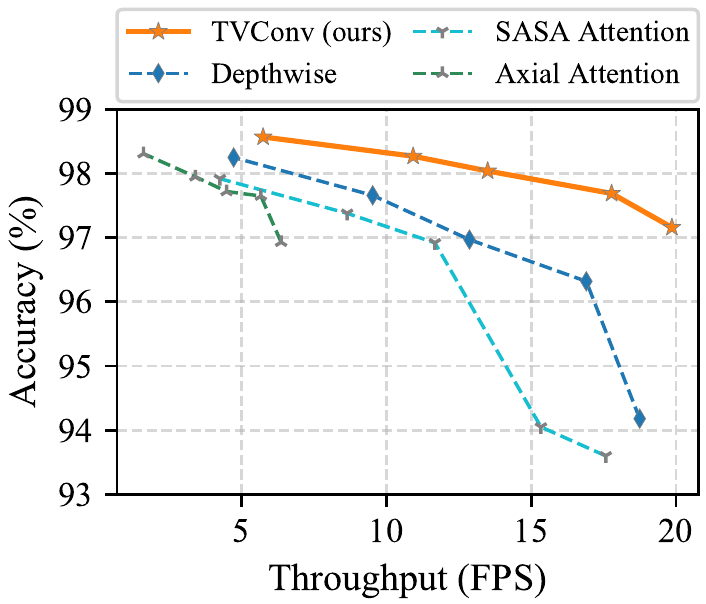}
        \caption{Against \textbf{self-attention} variants}
        \label{fig:AccThrMem_c}
    \end{subfigure}
    \begin{subfigure}[b]{\AccThrMemwidth \linewidth}
        \centering
        \includegraphics[width=\linewidth]{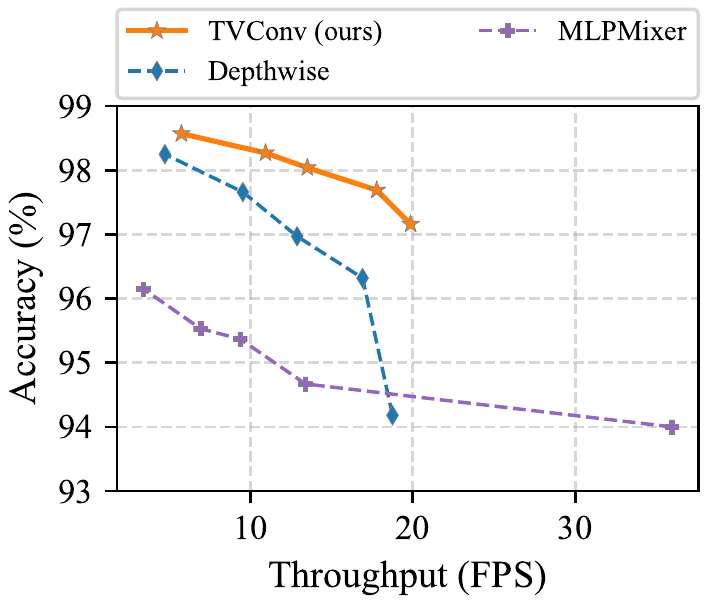}
        \caption{Against \textbf{MLPMixer}}
        \label{fig:AccThrMem_d}
    \end{subfigure}
    \vspace{-0.05in}
    \caption{TVConv achieves the best accuracy-throughput trade-off on the LFW dataset compared to related works.}
    \label{fig:AccThr}
    \vspace{-0.05in}
\end{figure*}

\begin{figure}
    \vspace{-0.2in}
    \newcommand \AccThrMemwidth{0.494}
    \centering
    \begin{subfigure}[b]{\AccThrMemwidth \linewidth}
        \centering
        \includegraphics[width=\linewidth]{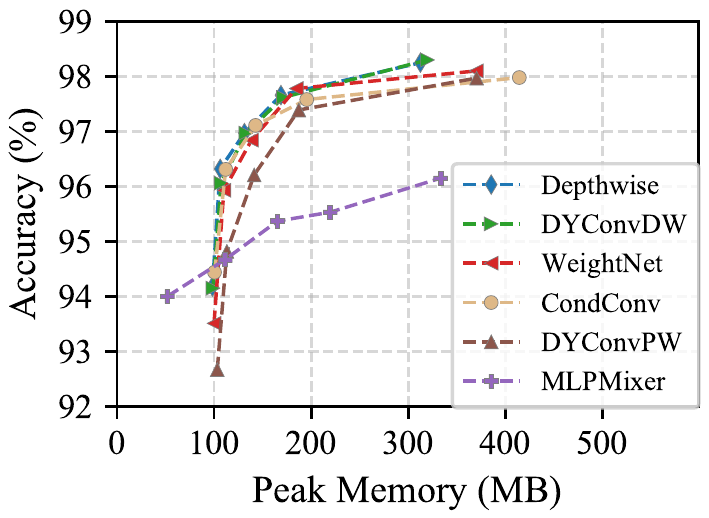}
        \caption{Highest peak memory $<$ 500MB}
        \label{fig:AccThrMem_e}
    \end{subfigure}
    \begin{subfigure}[b]{\AccThrMemwidth \linewidth}
        \centering
        \includegraphics[width=\linewidth]{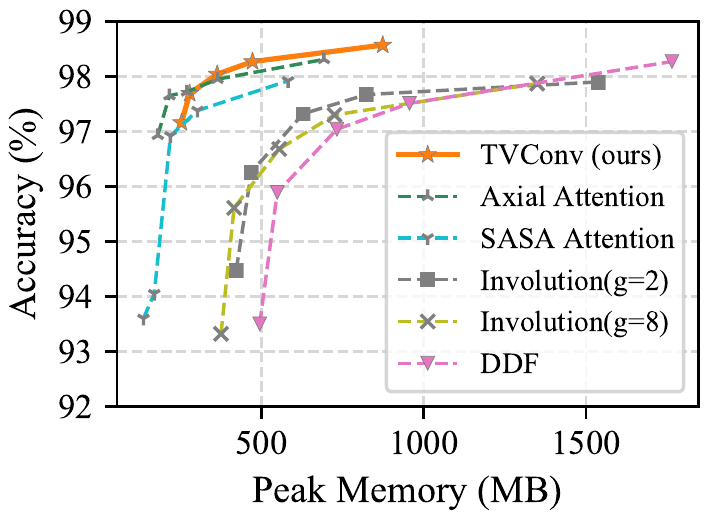}
        \caption{Highest peak memory $>$ 500MB}
        \label{fig:AccThrMem_f}
    \end{subfigure}
    \vspace{-0.1in}
    \caption{TVConv consumes a medium amount of peak memory.}
    \label{fig:AccMem}
    \vspace{-0.2in}
\end{figure}

\medskip\noindent\textbf{Comparison with depthwise conv.} \cref{fig:acc_comp_envelope} shows a clear gap between TVConv and the depthwise conv when applied on the MobileNetV2 (MB) and ShuffleNetV2 (SF) with varied network width. In particular, TVConv for MobileNetV2 reduces the complexity by 3.05x while maintaining a high accuracy on the LFW dataset. On the other hand, TVConv boosts the accuracy by 2.98\% under an extreme low complexity constraint. More quantitative comparisons can be found in \cref{tab:varied_width}, where a more significant 4.21\% mean accuracy boost of applying TVConv on MBx0.1 is achieved. Throughput analysis in \cref{fig:AccThr} shows a 2.3x improvement of MBx0.5 with TVConv over the original MBx1.0, though TVConv brings a fairly higher peak memory consumption as in shown \cref{fig:AccMem}. Note that the translation variant weights from TVConv reside on memory for query with only the fast READ operation performed.

\begin{figure*}
    \centering
    \begin{subfigure}[b]{.25\textwidth}
    \includegraphics[width=1\linewidth]{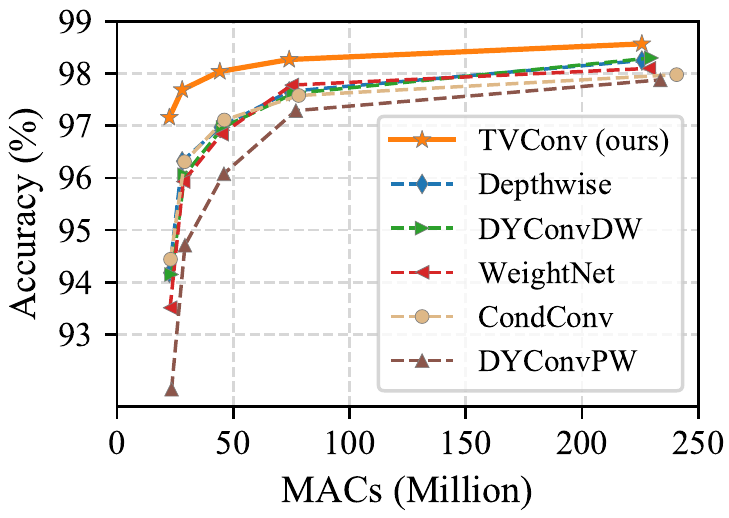}
    \caption{Against \textbf{per-image} dynamic conv}
  \end{subfigure}
  \begin{subfigure}[b]{.25\textwidth}
    \includegraphics[width=1\linewidth]{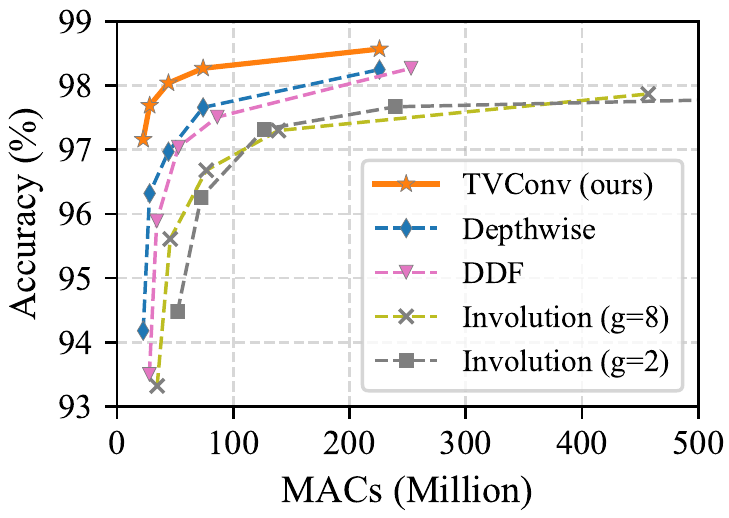}
    \caption{Against \textbf{per-pixel} dynamic conv}
  \end{subfigure}
  \begin{subfigure}[b]{.25\textwidth}
    \includegraphics[width=1\linewidth]{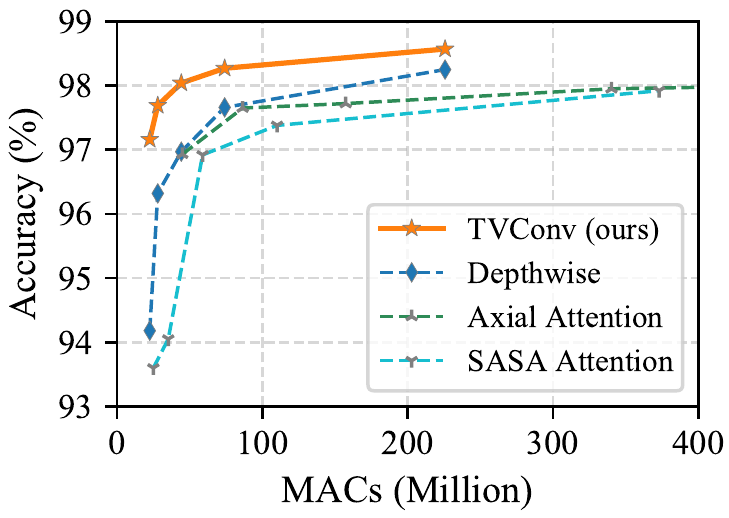}
    \caption{Against \textbf{self-attention} variants}
  \end{subfigure}
  \begin{subfigure}[b]{.236\textwidth}
    \includegraphics[width=1\linewidth]{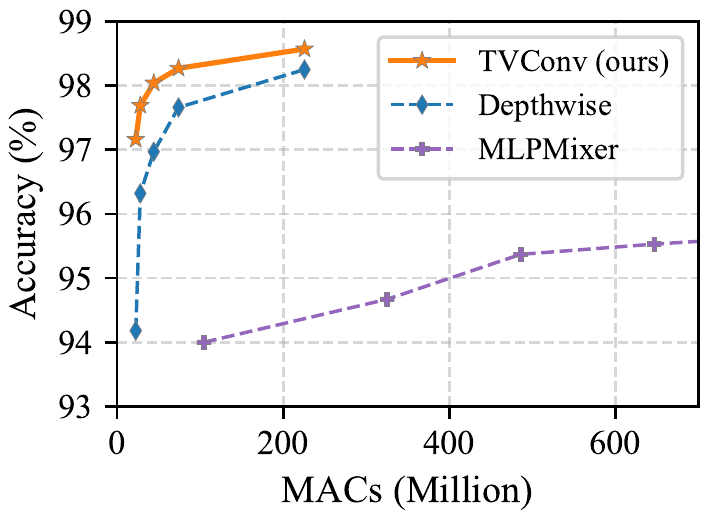}
    \caption{Against \textbf{MLPMixer}}
  \end{subfigure}
  \caption{TVConv strikes a better accuracy-complexity envelope on the LFW dataset compared to four related branches of works.}
  \label{fig:acc_macs_atoms}
  \vspace{-0.12in}
\end{figure*}

\vspace{-0.05in}

\medskip\noindent\textbf{Comparison with other operators.} We compare with the related per-image dynamic conv, per-pixel dynamic conv and self-attention variants applied on the MobileNetV2 as well as the standalone MLPMixer. In particular, we compare with CondConv~\cite{yang2019condconv}, WeightNet~\cite{ma2020weightnet}, DYConvDW~\cite{chen2020dynamic}, DYConvPW~\cite{chen2020dynamic} as the per-image dynamic conv variants and with Involution~\cite{li2021involution}, DDF~\cite{zhou2021decoupled} as the per-pixel dynamic conv variants. They are put at every stage of the network, as the way of TVConv being placed. SASA~\cite{ramachandran2019stand} and Axial attention~\cite{wang2020axial} are compared as two efficient self-attention variants, each put at the last three stages of the network for better efficiency. Note that the MLPMixer~\cite{tolstikhin2021mlp} is out of the conv family. We include it for comparison for its potential layout-aware ability. Results are summarised as three types of trade-offs: the accuracy-complexity trade-off (see \cref{fig:acc_macs_atoms}), accuracy-throughput trade-off  (see \cref{fig:AccThr}) and accuracy-peak memory trade-off (see \cref{fig:AccMem}). TVConv consistently outperforms other operators with a better efficiency on accuracy-complexity and accuracy-throughput trade-offs. The gap is remarkable, especially with small network width where the spatial adaptiveness becomes more crucial. Besides, we find that many works of per-image dynamic conv tend to clutter together on the accuracy-complexity trade-off without clear boundaries and perform even inferior to the depthwise conv. This is because the cross-image dynamics are useless here for face recognition with small cross-image variance. For per-pixel dynamic conv, the weight-generating process incurs an expensive computational overhead for marginal accuracy gain. Though the Axial attention greatly improves the accuracy over the depthwise baseline, the computational overhead is still unaffordable. For the MLPMixer, it can easily achieve ultra-fast speed (\eg 36.5 FPS for 94.00\% accuracy). However, the accuracy grows slowly as the model size scales up. In contrast, TVConv keeps its high accuracy even at its smallest version (19.8 FPS for 97.16\% accuracy). For peak memory usages, TVConv occupies less memory than the per-pixel dynamic conv and self-attention variants but fairly higher than others.

\medskip\noindent\textbf{Learned affinity maps visualization.} To further investigate the spatial adaptivity of TVConv, we visualize the learned affinity maps. As shown in the bottom row of \cref{fig:aff_visualization}, the affinity map progressively converges to a face contour. However, such a clear learning process does not occur for each layer. As in the top and middle rows of \cref{fig:aff_visualization}, the very early layers are mainly kept as a constant map without learning much meaningful layout information. This suggests that the very low-level features tend to be shared across the whole image probably because of the texture similarities and the presence of signal noise. For deeper layers, the learned affinity maps do not appear in the shape of a human face either. This aligns with the paper~\cite{simonyan2014very} saying that deeper layers are responsible for more abstract semantic concepts. Interestingly, we observe that for some affinity maps in deep layers, the most ``activated'' region does not appear at the center but rather tends to appear on the corners. We hypothesize that these corners carry more descriptive features. Because in the more peripheral region, its receptive filed covers a larger blank area, to which the zero-padding is applied. Such zero-padding would help exploit the spatial information, as indicated in the work~\cite{kayhan2020translation}.

\subsection{Optic disc/cup segmentation}
Due to the layout awareness, we expect this geometrical property to be helpful under a small data regime, particularly for medical image processing. To examine this, we conduct experiments on the optic disc and cup (OD, OC) segmentation tasks. We employ the fundus images from 4 clinical centers of public datasets~\cite{sivaswamy2015comprehensive, fumero2011rim,orlando2020refuge}, which are considered as data from different domains. Following the work~\cite{wang2020dofe}, we take turns to use data from three domains for training and the remaining domain for evaluation. DeepLabV3+~\cite{chen2018encoder} network is used as a strong baseline. For comparison, we replace the conv in its ASPP module with the inverted residual block~\cite{sandler2018mobilenetv2}, which is equipped with either the original depthwise conv or our TVConv as two variants. An Adam optimizer is used to train for 40 epochs. The learning rate starts from  $1\text{e-}3$ and is divided by 5 at epoch 30. Images are resized to 256 x 256 with batch size 16 for training but kept in the original size of 800 x 800 for evaluation. To make the data scarcity even more severe, we do not employ any data augmentation. For evaluation, we use two commonly adopted metrics, dice similarity coefficient (DSC) and Hausdorff distance (HD).

\begin{figure}
    \newcommand \medicalwidth{0.23}
    \centering
    \begin{subfigure}[b]{\medicalwidth \linewidth}
        \centering
        \includegraphics[width=\linewidth]{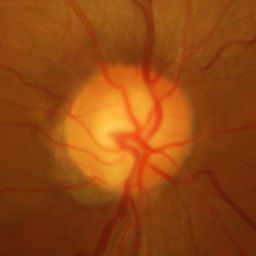}
    \end{subfigure}
    \begin{subfigure}[b]{\medicalwidth \linewidth}
        \centering
        \includegraphics[width=\linewidth]{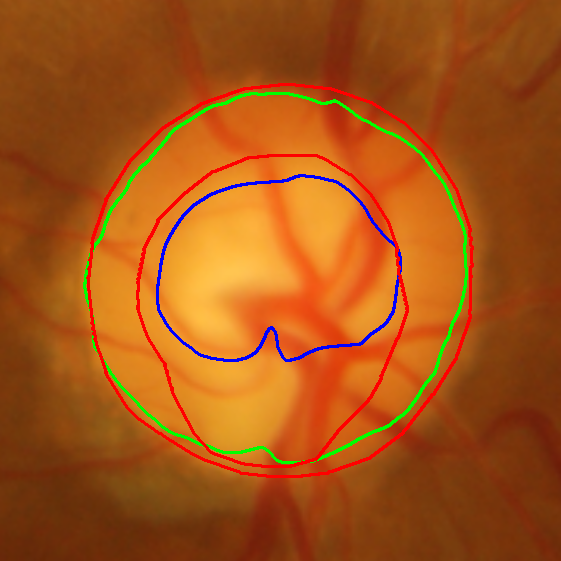}
    \end{subfigure}
    \begin{subfigure}[b]{\medicalwidth \linewidth}
        \centering
        \includegraphics[width=\linewidth]{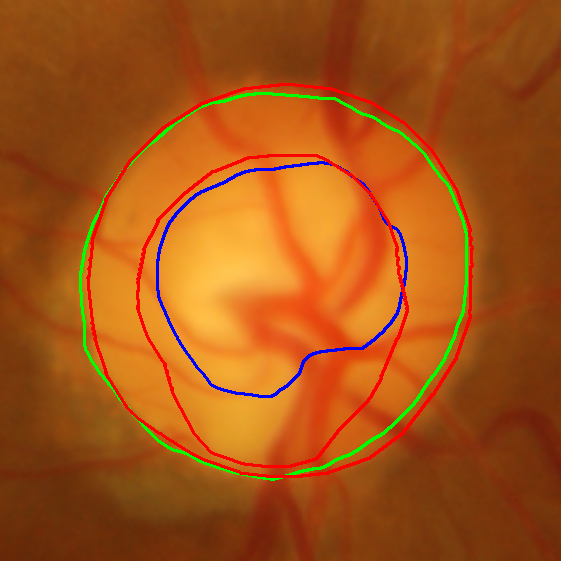}
    \end{subfigure}
    \begin{subfigure}[b]{\medicalwidth \linewidth}
        \centering
        \includegraphics[width=\linewidth]{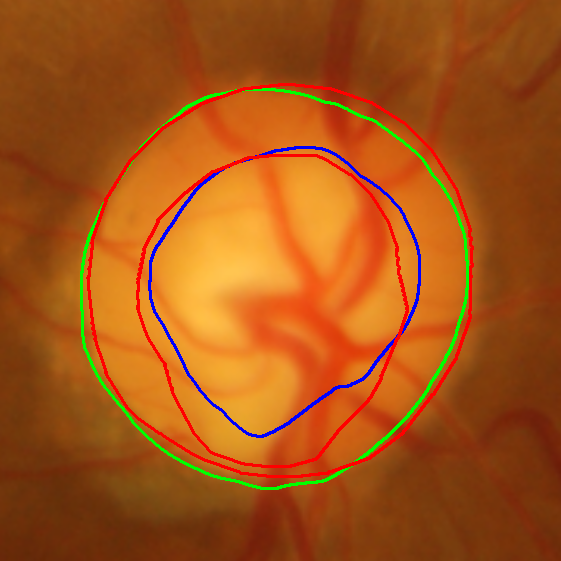}
    \end{subfigure}
    \hfill
    
    \begin{subfigure}[b]{\medicalwidth \linewidth}
        \centering
        \includegraphics[width=\linewidth]{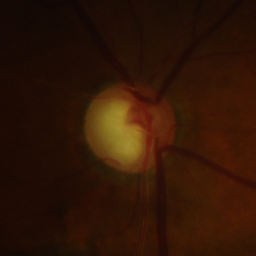}
    \end{subfigure}
    \begin{subfigure}[b]{\medicalwidth \linewidth}
        \centering
        \includegraphics[width=\linewidth]{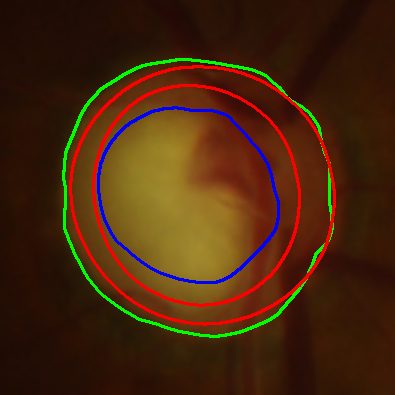}
    \end{subfigure}
    \begin{subfigure}[b]{\medicalwidth \linewidth}
        \centering
        \includegraphics[width=\linewidth]{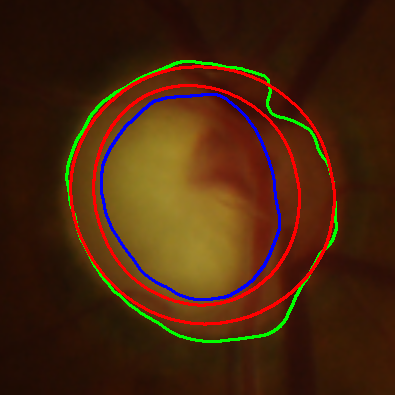}
    \end{subfigure}
    \begin{subfigure}[b]{\medicalwidth \linewidth}
        \centering
        \includegraphics[width=\linewidth]{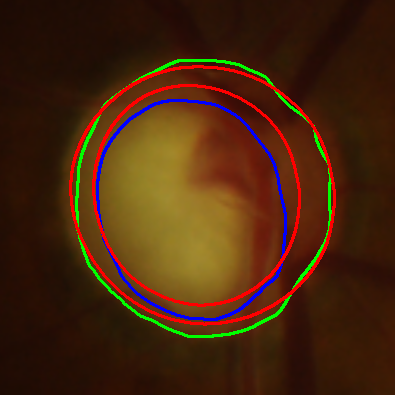}
    \end{subfigure}
    \hfill

    \begin{subfigure}[b]{\medicalwidth \linewidth}
        \centering
        \includegraphics[width=\linewidth]{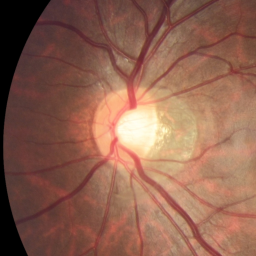}
        \caption{Inputs}
    \end{subfigure}
    \begin{subfigure}[b]{\medicalwidth \linewidth}
        \centering
        \includegraphics[width=\linewidth]{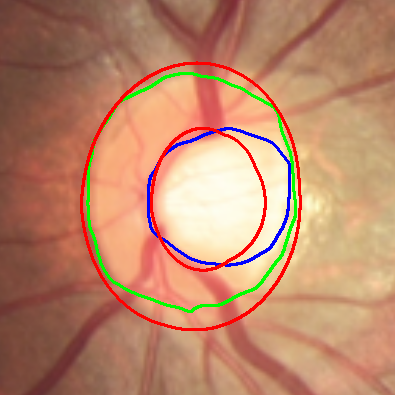}
        \caption{DeepLabV3+}
    \end{subfigure}
    \begin{subfigure}[b]{\medicalwidth \linewidth}
        \centering
        \includegraphics[width=\linewidth]{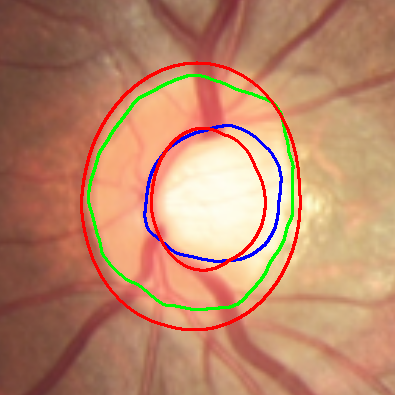}
        \caption{w/ Depthwise}
    \end{subfigure}
    \begin{subfigure}[b]{\medicalwidth \linewidth}
        \centering
        \includegraphics[width=\linewidth]{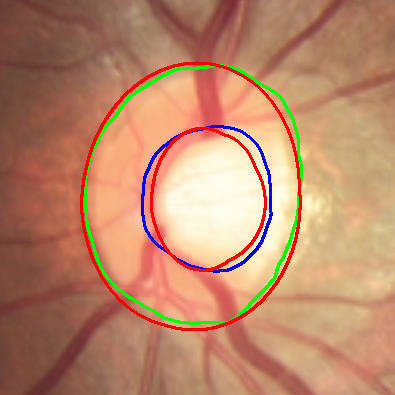}
        \caption{w/ TVConv}
    \end{subfigure}
    \hfill    
        
    \caption{Qualitative segmentation comparisons. Column (a) are the inputs while column (b), (c) and (d) show the amplified segmentation results using different approaches. The red contours represent the ground truth while the green and blue lines show the predictions for optic discs and cups, respectively.}
    \label{fig:medical}
    \vspace{-0.1in}
\end{figure}

As reported in \cref{tab:medical}, with our TVConv, the network surpasses the baseline and its variant (with depthwise conv) on most unseen domains. Notably, comparison with the depthwise conv is necessary to ensure that our performance gain mainly comes from the operator itself instead of the structural change. TVConv improves the baseline by a considerable margin (an average DSC of 1.52\%) compared to a slight improvement (an average DSC of 0.31\%) brought by the inverted residual blocks. Qualitative comparisons are also provided in \cref{fig:medical}. Our TVConv generates more accurate and resilient segmentation results while others tend to produce deviated curves in regions with locally complicated or blurry textures. This better generalization ability is desired and particularly useful in practice. Besides, unlike the sparse prediction tasks (\eg, face recognition), segmentation is in the category of dense prediction, indicating the broad applicability of TVConv in certain degree.
\begin{table}
\centering
\resizebox{.95\textwidth}{!}{%
\begin{tabular}{@{}cccccc@{}}
\toprule
Metric & Task & \begin{tabular}[c]{@{}c@{}}Unseen \\ domain\end{tabular} & DeepLabV3+ & w/ Depthwise & \cc{\; w/ TVConv\;\;} \\ \midrule
                      & \cellcolor[HTML]{FFFFFF}                     & A & 79.39 ± 2.83          & 78.34 ± 2.27          & \cc{\textbf{82.76 ± 1.41}} \\
                      & \cellcolor[HTML]{FFFFFF}                     & B & 75.85 ± 1.08          & 75.28 ± 1.34          & \cc{\textbf{76.21 ± 1.56}} \\
                      & \cellcolor[HTML]{FFFFFF}                     & C & \cc{\textbf{85.60 ± 0.61}} & 85.35 ± 0.79          & 85.52 ± 0.51          \\
                      & \multirow{-4}{*}{\cellcolor[HTML]{FFFFFF}OC} & D & 81.24 ± 1.70          & 82.81 ± 0.94          & \cc{\textbf{84.10 ± 0.76}} \\ \cmidrule(l){2-6} 
                      & \cellcolor[HTML]{FFFFFF}                     & A & 92.18 ± 0.64          & \cc{\textbf{94.21 ± 0.90}} & 93.67 ± 1.24          \\
                      & \cellcolor[HTML]{FFFFFF}                     & B & 88.66 ± 0.78          & 89.57 ± 1.64          & \cc{\textbf{90.10 ± 0.69}} \\
                      & \cellcolor[HTML]{FFFFFF}                     & C & 90.89 ± 1.46          & 90.26 ± 1.63          & \cc{\textbf{92.91 ± 0.59}} \\
                      & \multirow{-4}{*}{\cellcolor[HTML]{FFFFFF}OD} & D & 92.52 ± 0.24          & 92.96 ± 0.51          & \cc{\textbf{93.25 ± 0.60}} \\ \cmidrule(l){2-6} 
\multirow{-9}{*}{DSC$\uparrow$} & \multicolumn{2}{c}{Mean}                                & 85.79                 & 86.10                  & \cc{\textbf{87.31}}        \\ \midrule
                      & \cellcolor[HTML]{FFFFFF}                     & A & 43.22 ± 3.86          & 42.65 ± 3.21          & \cc{\textbf{36.32 ± 2.79}} \\
                      & \cellcolor[HTML]{FFFFFF}                     & B & \cc{\textbf{30.69 ± 1.02}} & 31.75 ± 1.68          & 31.01 ± 1.46          \\
                      & \cellcolor[HTML]{FFFFFF}                     & C & \cc{\textbf{21.44 ± 0.76}} & 21.85 ± 1.10          & 21.71 ± 0.78          \\
                      & \multirow{-4}{*}{\cellcolor[HTML]{FFFFFF}OC} & D & 21.17 ± 1.56          & 19.56 ± 1.32          & \cc{\textbf{17.97 ± 1.14}} \\ \cmidrule(l){2-6} 
                      & \cellcolor[HTML]{FFFFFF}                     & A & 25.23 ± 1.22          & \cc{\textbf{20.36 ± 2.18}} & 21.96 ± 3.82          \\
                      & \cellcolor[HTML]{FFFFFF}                     & B & 31.63 ± 2.82          & 29.95 ± 3.21          & \cc{\textbf{27.69 ± 3.01}} \\
                      & \cellcolor[HTML]{FFFFFF}                     & C & 24.62 ± 3.53          & 26.17 ± 3.75          & \cc{\textbf{20.88 ± 1.55}} \\
                      & \multirow{-4}{*}{\cellcolor[HTML]{FFFFFF}OD} & D & 19.34 ± 0.94          & 17.75 ± 1.25          & \cc{\textbf{17.05 ± 1.14}} \\ \cmidrule(l){2-6} 
\multirow{-9}{*}{HD$\downarrow$}  & \multicolumn{2}{c}{Mean}                                & 27.17                 & 26.26                 & \cc{\textbf{24.32}}        \\ \bottomrule
\end{tabular}%
}
\caption{Comparison of domain generalization results on OD/OC segmentation tasks. Ten runs for each setting.}
\label{tab:medical}
\vspace{-0.05in}
\end{table}

\subsection{Ablation study}\label{ablation}
We conduct experiments to answer the following four questions. Unless otherwise specified, all experiments use a MBx0.2 architecture and are evaluated on the LFW dataset.

\begin{figure}
    \vspace{-0.15in}
    \centering
    \begin{subfigure}[b]{0.45\linewidth}
        \centering
        \includegraphics[width=\linewidth]{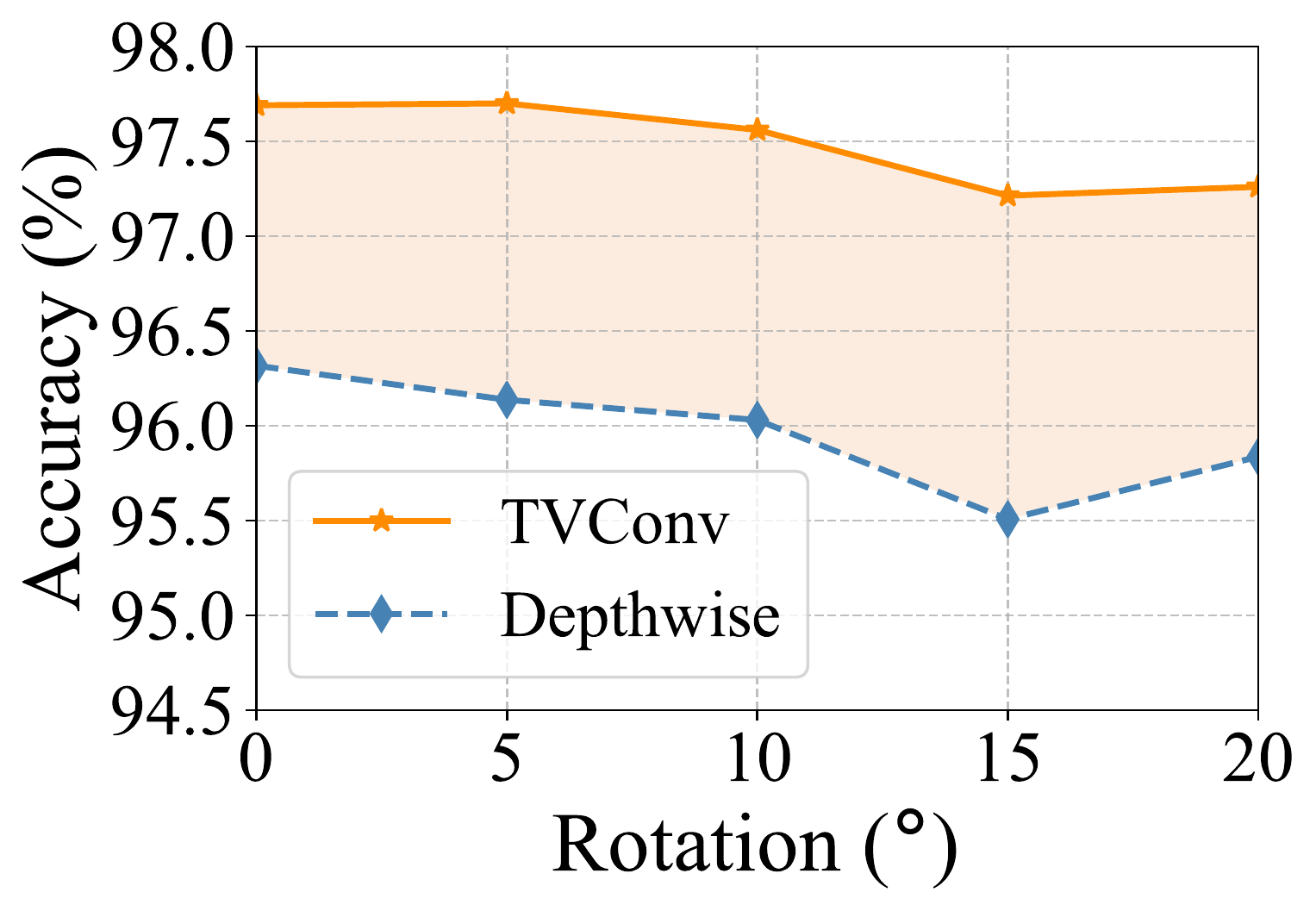}
    \end{subfigure}
    \begin{subfigure}[b]{0.45\linewidth}
        \centering
        \includegraphics[width=\linewidth]{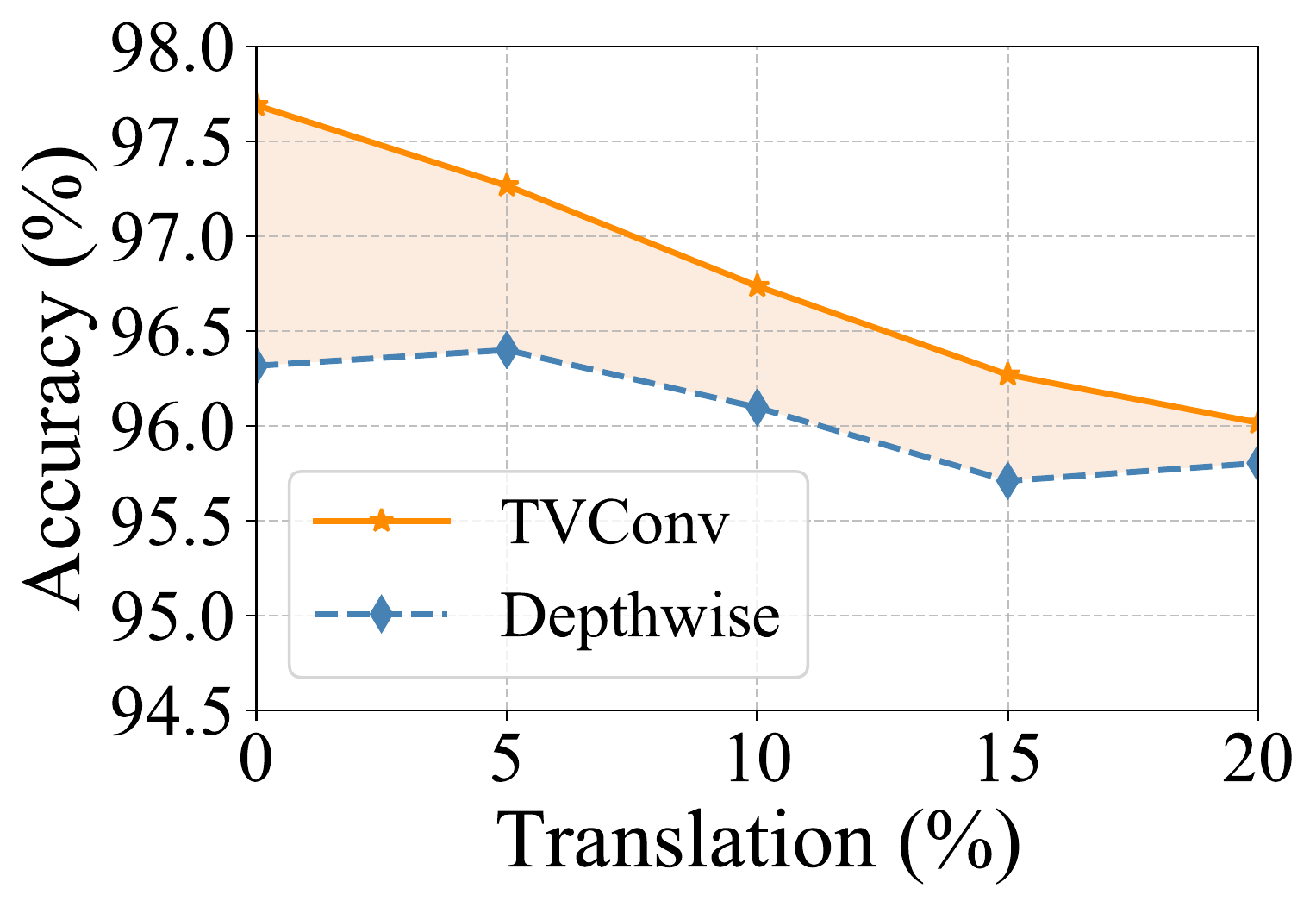}
    \end{subfigure}
    \begin{subfigure}[b]{0.45\linewidth}
        \centering
        \includegraphics[width=\linewidth]{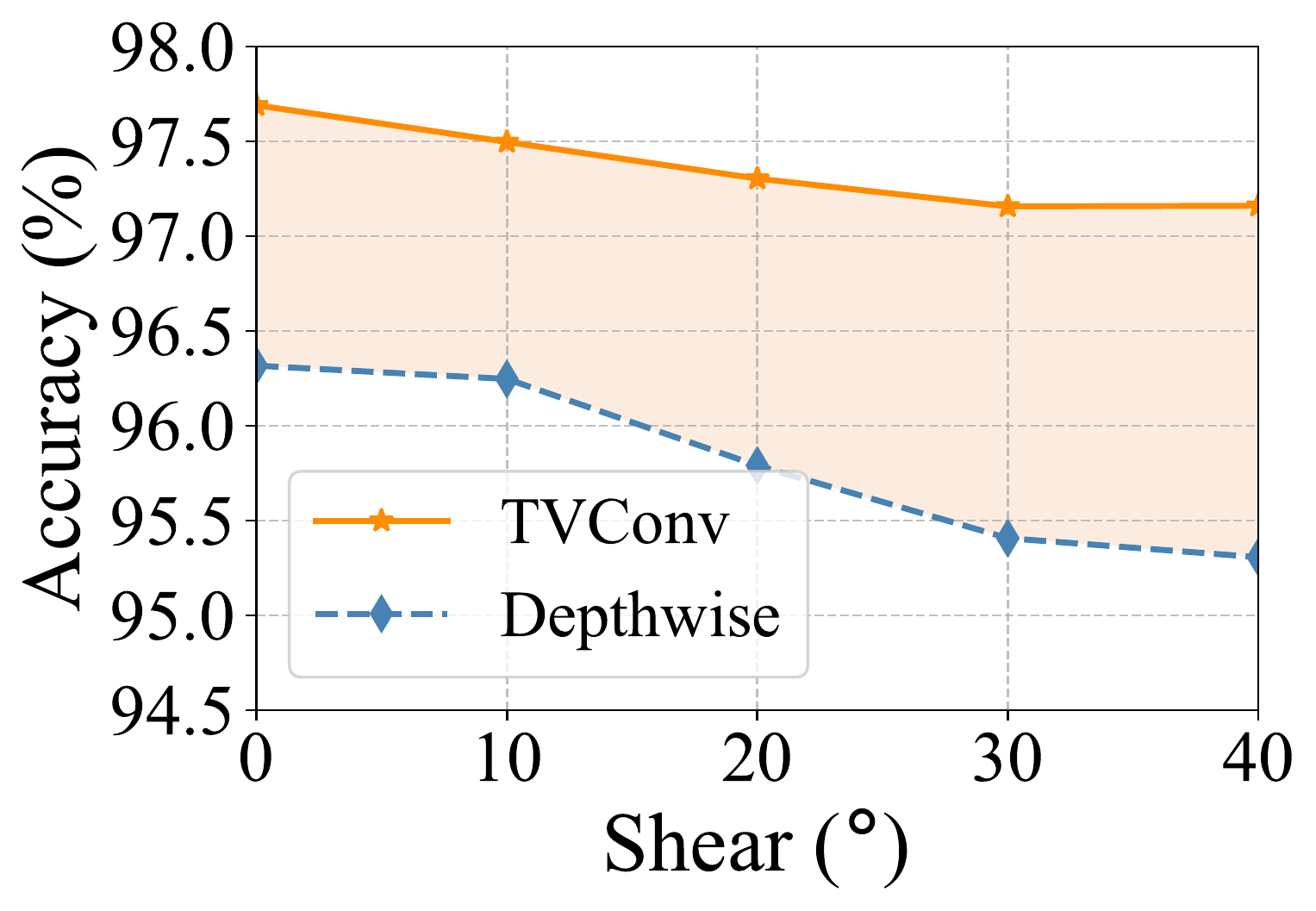}
    \end{subfigure}
    \begin{subfigure}[b]{0.45\linewidth}
        \centering
        \includegraphics[width=\linewidth]{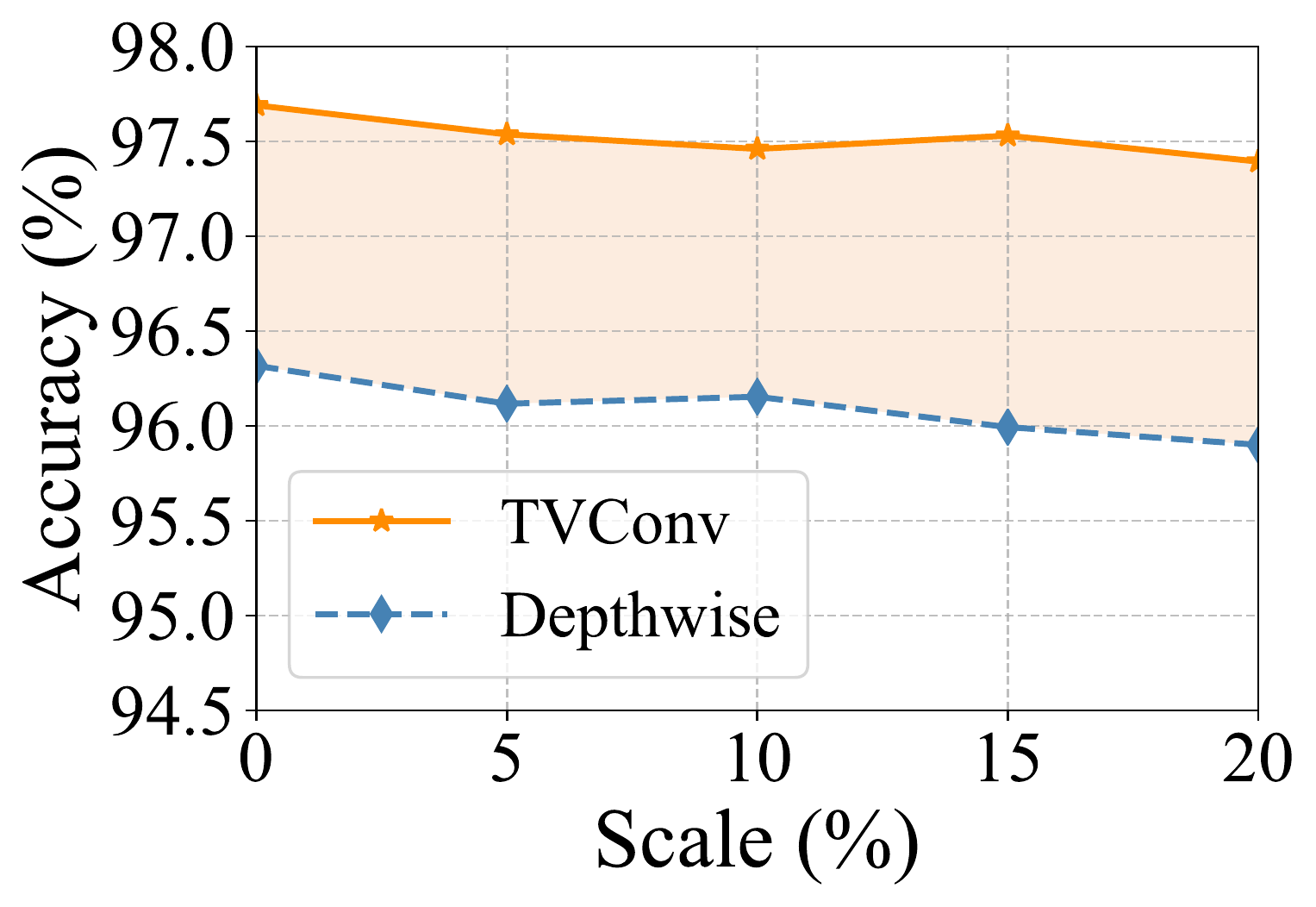}
    \end{subfigure}
    \caption{TVConv surpasses the depthwise conv for various transformations, while for translation the gap gradually shrinks.}
    \label{fig:sensitivity}
    \vspace{-0.15in}
\end{figure}

\medskip\noindent\textbf{How sensitive is TVConv to various affine transformations?} For some applications (\eg face recognition for phone unlocking), it generally holds that the input image has a specific and fixed layout pattern. If not, an alignment would be involved and serve as a prepossessing step. Here, we explicitly relax this condition to investigate the sensitivity of TVConv. We apply four affine transforms independently to the training dataset and report the validation results in \cref{fig:sensitivity}. We can see that TVConv performs robustly under a reasonable amount of rotation, shearing, or scaling. It performs even more robustly than the depthwise conv against the shearing. But for translation, as expected, the accuracy gain gradually shrinks, and TVConv would degrade to a vanilla depthwise conv under severe translation.

\begin{figure}
    \centering
    \vspace{-0.1in}
    \hspace{0.02\textwidth}
    \begin{subfigure}[b]{0.18\linewidth}
        \centering
        \includegraphics[width=\linewidth]{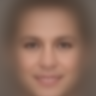}
        \caption{Mean}
    \end{subfigure}
    \hfill
    \begin{subfigure}[b]{0.18\linewidth}
        \centering
        \includegraphics[width=\linewidth]{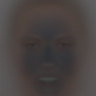}
        \caption{Std}
    \end{subfigure}
    \hfill
    \begin{subfigure}[b]{0.18\linewidth}
        \centering
        \includegraphics[width=\linewidth]{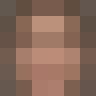}
        \caption{$D$(mean)}
    \end{subfigure}
    \hfill
    \begin{subfigure}[b]{0.18\linewidth}
        \centering
        \includegraphics[width=\linewidth]{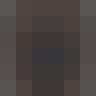}
        \caption{$D$(std)}
    \end{subfigure}
    \hfill
    \begin{subfigure}[b]{0.18\linewidth}
        \centering
        \includegraphics[width=\linewidth]{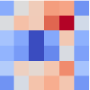}
        \caption{Learned}
    \end{subfigure}
\hspace{0.02\textwidth}
    \caption{Statistics calculated from the datasets: (a) mean, (b) std, as well as their down-sampled version (c), (d), respectively. These maps might be used to initialize the affinity maps. However, it may geometrically misalign with learned affinity maps (e).}
    \label{fig:ablation_ini}
    \vspace{-0.1in}
\end{figure}

\medskip\noindent\textbf{How should we initialize the affinity maps?}
By default, we initialize the affinity maps $\mathbf{A}$ to be constant maps (\eg, a constant 1). In this way, the training stability can be better guaranteed. Because the weight-generating block $\mathcal{B}$ produces identical weights for different positions at the beginning, and TVConv starts as a plain depthwise conv. This warm-up process enables TVConv to gradually evolve and manifest its adaptivity. Another possible way (see \cref{fig:ablation_ini}) is to initialize from the data statistics, \eg to calculate the mean and std maps from the training images, and then down-sample them to a specific size identical to the affinity maps. Though it seems promising, our experiments (see \cref{tab:ablation_ini}) suggest that it does not provide better performance than simple constant initialization. Probably because in the post layers of a network, the learned affinity maps generally carry some high-level semantic information, which can be visually misaligned with the input contours. 

\medskip\noindent\textbf{Does the model benefit from a bigger weight-generating block?} In line with the work~\cite{allen2019convergence}, over-parameterization generally brings better performance in our experiments, as in \cref{tab:abltaion_generating_block}. In particular, we examine the role of different components, including the depth (layers), width (channels) of the weight-generating block $\mathcal{B}$ and the number of channels in affinity maps $\mathbf{A}$. With deeper, wider layers in $\mathcal{B}$ or more channels in $\mathbf{A}$, the model generally performs better. However, the performance slightly drops from the peak when the over-parameterization goes too far. This is because a larger model might require more iterations for training and can suffer from over-fitting.

\medskip\noindent\textbf{Where should TVConv be placed in a network?}
Different patterns appeared in the learned affinity maps (see \cref{fig:aff_visualization}) motivate us to further investigate the proper position for placing our TVConv. As reported in \cref{tab:ablation_stage}, applying a single TVConv in post stages is more effective in reducing the error while applying in early stages (\eg S1, S2) might even hurt the performance. Moreover, by stacking more stages with TVConv, the error is monotonically reduced. We place TVConv in all stages as our default setting.
\begin{table}
\centering
\vspace{-0.05in}
\resizebox{.95\linewidth}{!}{%
\begin{tabular}{@{}cccc@{}}
\toprule
Init. & Error (\%) for ×0.1  & Error (\%) for ×0.5  & Error (\%) for ×1.0  \\ \midrule
data statistics          & 2.92 ± 0.22          & 1.75 ± 0.18          & 1.62 ± 0.20          \\
\cc{\textbf{constant 1}}             & \cc{\textbf{2.84 ± 0.11}} & \cc{\textbf{1.73 ± 0.07}} & \cc{\textbf{1.43 ± 0.12}} \\ \bottomrule
\end{tabular}%
}
\caption{Initialization from constant maps (\eg constant 1) works better than initialization from the data statistics.}
\label{tab:ablation_ini}
\vspace{-0.08in}
\end{table}
\begin{table}
\centering
\vspace{-0.1in}
\resizebox{\linewidth}{!}{%
\begin{tabular}{@{}cc|cc|cc@{}}
\toprule
$\mathcal{B}$ \#Chans. &
  Error (\%)  &
  $\mathcal{B}$ \#Layers &
  Error (\%)  &
  $\mathbf{A}$ \#Chans. &
  Error (\%)  \\ \midrule
128 & 2.60 ± 0.10  & 4 & 2.49 ± 0.16 & 8 & 2.49 ± 0.16 \\
\cc{\textbf{64}} & \cc{\textbf{2.31 ± 0.12}} & \cc{\textbf{3}} & \cc{\textbf{2.31 ± 0.12}} & \cc{\textbf{4}} & \cc{\textbf{2.31 ± 0.12}}  \\
32  & 2.65 ± 0.21 & 2 & 2.44 ± 0.15 & 2 & 2.44 ± 0.19 \\
8   & 2.65 ± 0.12 & 1 & 2.48 ± 0.27& 1 & 2.52 ± 0.15  \\ \bottomrule
\end{tabular}%
}
\caption{Ablation under the varied number of channels and inter layers for weight generating block $\mathcal{B}$, as well as the varied number of channels for affinity maps $\mathbf{A}$. We use hyper-parameters highlighted in bold as our default setting elsewhere in the paper.}
\label{tab:abltaion_generating_block}
\vspace{-0.05in}
\end{table}

\begin{table}
\vspace{-0.05in}
\resizebox{\linewidth}{!}{
\begin{tabular}{c|ccccccc}
\toprule
Stages & S1  & S2 & S3& S4 & S5 & S6 & S7 \\
Error(\%) & 4.01 & 3.73         & 3.42 & 3.43 & 3.34 & 3.25 & 3.39 \\ \hline
Stages & Baseline & \cc{\textbf{S1-S7}}     & S2-S7     & S3-S7      & S4-S7       & S5-S7       & S6-S7   \\
Error(\%) & 3.68 & \cc{\textbf{2.31}} & 2.35& 2.38 & 2.51 & 2.53 & 2.61 \\ \bottomrule
\end{tabular}
}
\caption{Ablation of applying TVConv in different stages. Excluding the input layer, MobileNetV2 has 7 stages. Stages S1-S7 correspond to layers L1, L2-L3, L4-L6, L7-L10, L11-L13, L14-L16 and L17, respectively. Error SD is omitted for simplicity.}
\label{tab:ablation_stage}
\vspace{-0.08in}
\end{table}


\section{Conclusion and Future Works}
\label{sec:conclusion}

We present a conceptually simple, novel, and efficient fundamental operator TVConv as a versatile replacement of existing convolution variants for layout-aware visual processing. A comprehensive study confirms the promising efficiency as well as the better generalization ability. TVConv can be seamlessly integrated into various neural architectures. Moreover, the learned affinity maps provide additional space for network visualization and interpretation. Nonetheless, TVConv has its own limitations. Since our prototype assumes the affinity maps to be of fixed size for a certain layer, networks with TVConv might fail to process inputs of varying resolutions. This might be addressed by adaptive pooling or sampling techniques. Besides, TVConv may degrade to a plain depthwise convolution when faced with severe layout transformations. One intuitive idea to solve this is to work with a simple Spatial Transformer Network~\cite{jaderberg2015spatial}, which produces a unified affine transformation matrix to globally transform the affinity maps in different layers. Another promising extension is to work with the dynamic convolution in a complementary way by efficiently producing pixel-wise dynamic bias. These two extensions would broaden the reach of our TVConv to serve more unconstrained scenarios, which are left for future work.
\section*{Acknowledgement}
This work was supported, in part, by Hong Kong General Research Fund (under grant number 16200120).

{\small
\bibliographystyle{ieee_fullname}
\bibliography{6_egbib}
}

\end{document}